\documentclass[sigconf]{acmart}

\usepackage{parskip}
\usepackage{amsmath}
\usepackage{graphicx}
\usepackage{multirow}
\usepackage{float}
\usepackage{tensor}
\usepackage{bm}
\usepackage{caption}
\AtBeginDocument{%
  }

\copyrightyear{2025}
\acmYear{2025}
\setcopyright{acmlicensed}\acmConference[MM '25]{Proceedings of the 33rd ACM International Conference on Multimedia}{October 27--31, 2025}{Dublin, Ireland}
\acmBooktitle{Proceedings of the 33rd ACM International Conference on Multimedia (MM '25), October 27--31, 2025, Dublin, Ireland}
\acmDOI{10.1145/3746027.3755461}
\acmISBN{979-8-4007-2035-2/2025/10}

\acmConference[MM '25]{Make sure to enter the correct
  conference title from your rights confirmation email}{October 27--31,
  2025}{Dublin, Ireland}

\makeatletter
\DeclareMathSizes{5}{5}{4}{3}
\makeatother

\newcommand{\myPara}[1]{\vspace{0pt}\noindent\textbf{#1}}

\AtEndPreamble{
    \usepackage[capitalize]{cleveref}
    \crefname{section}{Sec.}{Secs.}
    \Crefname{section}{Section}{Sections}
    \Crefname{table}{Table}{Tables}
    \crefname{table}{Tab.}{Tabs.}
}

\def\figref{\cref}
\def\tabref{\cref}

\begin{document}

\title{Synthetic-to-Real Camouflaged Object Detection}

\author{Zhihao Luo}
\orcid{0009-0002-7424-0085}
\affiliation{%
 \institution{Fuzhou University}
 \city{Fuzhou}
 \country{China}}
\email{zhluo2003@gmail.com}

\author{Luojun Lin}
\orcid{0000-0002-1141-2487}
\authornote{L. Lin and Z. Lin are corresponding authors.}
\affiliation{%
  \institution{Fuzhou University}
  \city{Fuzhou}
  \country{China}
}
\email{linluojun2009@126.com}

\author{Zheng Lin}
\orcid{0000-0002-8057-4949}
\authornotemark[1]
\affiliation{%
  \institution{Tsinghua University}
  \city{Beijing}
  \country{China}
}
\email{frazer.linzheng@gmail.com}

\renewcommand{\shortauthors}{Zhihao Luo, Luojun Lin, \& Zheng Lin}

\begin{abstract}
  Due to the high cost of collection and labeling, there are relatively few datasets for camouflaged object detection (COD). In particular, for certain specialized categories, the available image dataset is insufficiently populated. Synthetic datasets can be utilized to alleviate the problem of limited data to some extent. However, directly training with synthetic datasets compared to real datasets can lead to a degradation in model performance. To tackle this problem, in this work, we investigate a new task, namely Syn-to-Real Camouflaged Object Detection (S2R-COD). In order to improve the model performance in real world scenarios, a set of annotated synthetic camouflaged images and a limited number of unannotated real images must be utilized. We propose the Cycling Syn-to-Real Domain Adaptation Framework (CSRDA), a method based on the student-teacher model. Specially, CSRDA propagates class information from the labeled source domain to the unlabeled target domain through pseudo labeling combined with consistency regularization. Considering that narrowing the intra-domain gap can improve the quality of pseudo labeling, CSRDA utilizes a recurrent learning framework to build an evolving real domain for bridging the source and target domain. Extensive experiments demonstrate the effectiveness of our framework, mitigating the problem of limited data and handcraft annotations in COD. Our code is publicly available at \href{https://github.com/Muscape/S2R-COD}{https://github.com/Muscape/S2R-COD}.
\end{abstract}

\ccsdesc[500]{Computing methodologies~Scene understanding}
\ccsdesc[500]{Computing methodologies~Image segmentation}

\keywords{Camouflaged Object Detection; Synthetic-to-Real; Noisy Label Learning}

\maketitle

\section{Introduction}
\label{sec:intro}
Camouflaged object detection (COD) focuses on identifying objects that blend seamlessly into their surroundings, characterized by low contrast, blurred boundaries, and a high degree of similarity between the object and the background \cite{MirrorNet, MGL, MGL-V2, SENet, ERRNet}. These features make COD significantly more challenging than standard object detection tasks. While substantial progress has been made in recent years \cite{camofocus, zoom, FEMNet, MLKG, RefCOD}, the complexity of real world scenes has rendered handcraft annotation both time-consuming and costly. For instance, annotating a single image in COD10K dataset takes approximately one hour \cite{SINet}, which limits the availability of datasets and restricts their diversity to a small range of species.

\begin{figure}[t]
    \centering
    \includegraphics[width=1\linewidth]{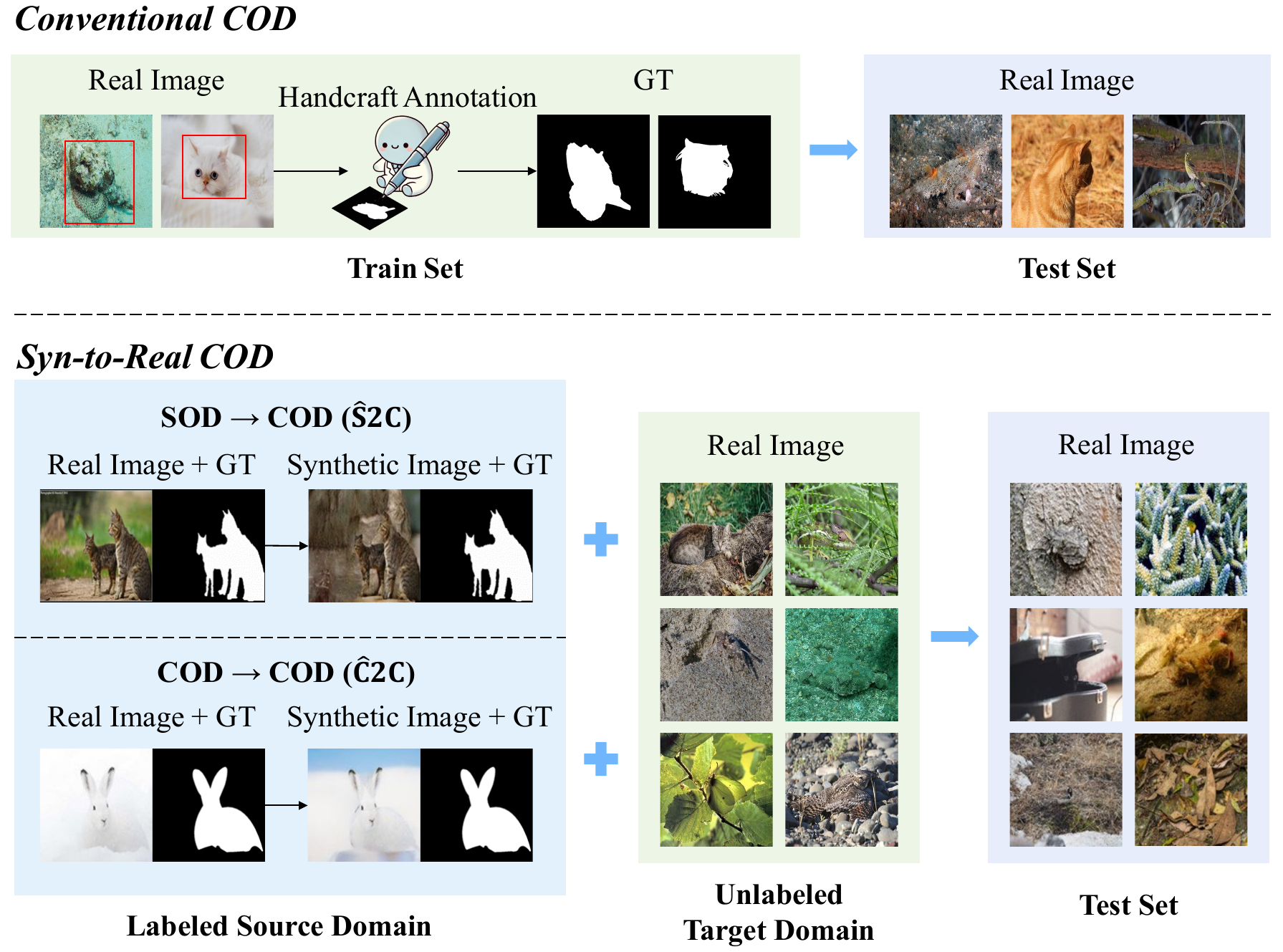}
    \caption{The difference between Conventional COD and Syn-to-Real COD. In conventional COD tasks, annotated data is scarce due to the difficulty of obtaining camouflaged objects, and handcraft annotation is costly, but Syn-to-Real COD approach utilizes synthetic data for supervised training and adapts to real, unlabeled images.}
    \vspace{-10pt}
    \label{fig:1}
\end{figure}

\begin{figure}[t]
    \centering
    \includegraphics[width=1\linewidth]{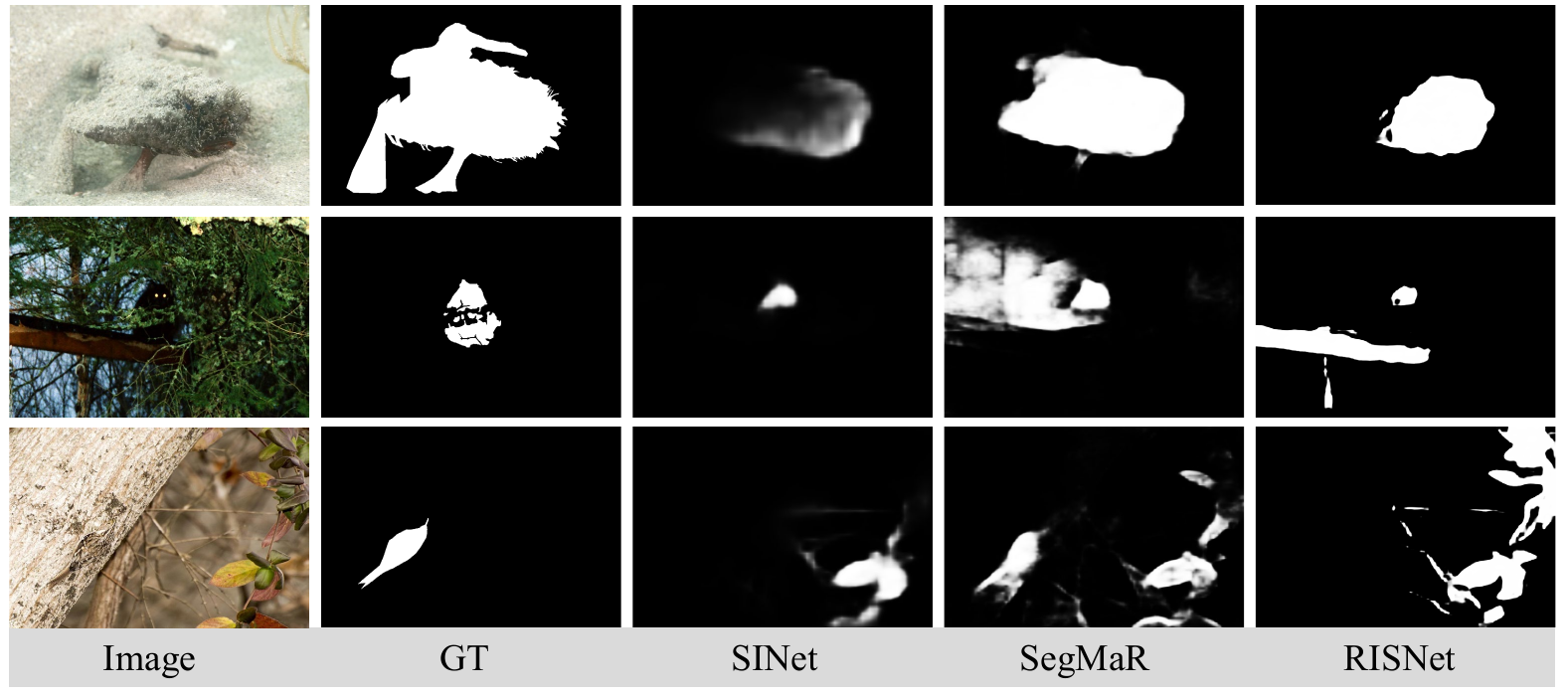}
    \caption{The output results of different models on the test set when trained using only source domain (synthetic images).}
    \label{fig:2}
    \vspace{-10pt}
\end{figure}

\hspace{1em}Data scarcity and the high cost of handcrafted annotation are common problems in many research areas. Using synthetic images to replace real data has become an effective and widely adopted strategy. In recent years, various methods for generating synthetic images have been proposed, such as LAKE-RED \cite{lake}, a method was proposed to generate synthetic camouflaged object images, with the aim of addressing the scarcity of real datasets and the limited number of images for certain species. Although the generated data set partially alleviates these issues, directly training models on synthetic images often degrades performance. A comparison between synthetic and real images reveals significant shortcomings in the synthetic data. These images appear designed solely to achieve the goal of blending objects into backgrounds, overlooking the realistic existence of objects. For example, as shown in \figref{fig:1}, the rabbit appears to be placed against a sky background, and the cats look unnaturally embedded in the rock. These scenarios are highly improbable under real world conditions. Furthermore, as illustrated in \figref{fig:2}, even the advanced model RISNet \cite{RISNet} performs poorly when trained solely on synthetic source domain images.

\hspace{1em}To address this limitation, we introduce the concept of Unsupervised Domain Adaptation (UDA). UDA is an effective approach to address the distribution shift between the source and target domain \cite{analysis}, particularly in scenarios where labeled data is scarce. In most applications, the training dataset is usually derived from the source domain, while the target domain is the scenario in which the model is actually applied. Therefore, we propose a novel task in this paper: Syn-to-Real Camouflaged Object Detection (S2R-COD). As shown in \figref{fig:1}, we compare conventional COD with our proposed S2R-COD. Categorizing tasks into \textbf{\boldmath$\hat{S}2C$} and {\textbf{\boldmath$\hat{C}2C$}}, \textbf{\boldmath$\hat{S}2C$} denotes using SOD dataset synthetic camouflaged dataset as the source domain. {\textbf{\boldmath$\hat{C}2C$}} indicates utilization of COD dataset as a synthetic camouflaged dataset for the source domain, noting that the source and target domain data do not overlap. The model is ultimately evaluated on the test set of real images. By effectively adapting from the synthetic domain to the real domain, this task provides a pathway to enhance the generalization capability of COD models, ultimately improving their performance in real world scenarios.

\hspace{1em}To effectively tackle the challenges posed by the S2R-COD task, we introduce a novel framework named Cycling Syn-to-Real Domain Adaptation Framework (CSRDA) that encourages pseudo labels generation and domain adaptation training to mutually benefit from a curriculum learning approach. The training process consists of two main stages: first, in inter-domain transfer phase, the student model is trained on the source domain data using supervised learning, while the target domain data is utilized in a consistency based unsupervised approach. To enhance adaptation, we introduce a customized Edge-Aware Saliency-Weighted Loss (ES Loss), which preserves edge structures and emphasizes salient regions by dynamically weighting pseudo labels. In this process, the teacher model is updated using the exponential moving average (EMA) \cite{meanteacher} of the student model. Then, in intra-domain transfer phase, we fix the trained student and teacher models and use them to generate pseudo labels for the target domain. These pseudo labels are filtered based on Confident Label Selection (CLS) mechanism and then bridged with the source domain data to construct an evolving real domain for the next training iteration. This process gradually reduces the domain gap and improves pseudo label quality.

\hspace{1em}We use SINet \cite{SINet} as the baseline model and train it with the proposed method, which is further extended to several classic COD models, such as SINet-v2 \cite{SINet-v2}, SegMaR \cite{SegMaR}, and RISNet \cite{RISNet}. We conducted extensive experiments on multiple datasets, including HKU-IS \cite{HKU-IS}, CAMO \cite{CAMO}, CHAMELEON \cite{CHAMELEON}, NC4K \cite{NC4K}, and COD10K \cite{SINet}, evaluating the effectiveness of the proposed method. The results show that the model trained with our approach achieves significant performance improvements in the target domain while maintaining stable performance across different datasets, demonstrating its effectiveness and broad applicability.

\hspace{1em}Overall, our contributions are summarized as follows:
\begin{itemize}
    \item We introduce the Syn-to-Real Camouflaged Object Detection (S2R-COD) task, establishing a new benchmark where synthetic data serves as the source domain and unlabeled real world data as the target domain.
\end{itemize}

\begin{itemize}
    \item We provide extensive strong benchmarks for S2R-COD by systematically evaluating traditional domain adaptation methods on multiple classic COD models.
\end{itemize}

\begin{itemize}
    \item We propose a Cycling Syn-to-Real Domain Adaptation Framework (CSRDA) based on student-teacher model. To boost domain adaptation, we design an Edge-Aware Saliency-Weighted Loss (ES Loss) and a Confident Label Selection (CLS) mechanism to align source and target domain progressively. Extensive experiments validate its effectiveness.
\end{itemize}

\section{Related Works}
\subsection{Camouflaged Object Detection}
Camouflaged Object Detection (COD) is a challenging computer vision task that focuses on identifying objects intentionally blended into their surroundings \cite{UR-COD, FS-CDIS, camofocus, VSCode, UJSCOD, ASBI}. The development of COD can be broadly divided into two stages: the traditional phase and the deep learning phase \cite{CMNet, BBNet, PAD, toward, FAP-Net, DCNet, PENet}. Traditional methods primarily used handcrafted features to differentiate camouflaged objects from their background, including texture \cite{texture}, intensity \cite{intensity}, and color \cite{color}. These features were used to compute attributes such as 3D convexity \cite{3Dconvexity}, optical flow \cite{opticalflow}, etc. These traditional methods perform well in simple scenarios but struggle with more complex cases. Deep learning has significantly advanced COD by enabling models to learn complex and high-level representations from data \cite{C2FNet, WS-SAM, HitNet, PFNet, POCINet}. Existing deep learning based COD methods can be broadly categorized into CNN-based and Transformer-based approaches. \textbf{a) CNN-based}: SINet \cite{SINet} uses the search and identification strategy to replicate the predation process of animals, achieving segmentation. SINet-v2 \cite{SINet-v2} proposes a neighbor connection decoder to refine feature representations, and group-reversal attention improves coarse predictions. DGNet \cite{DGNet} introduces the deep gradient network, which utilizes gradient-induced transitions to automatically group features. MFFN \cite{MFFN} employs Co-attention of multi-view to mine the complementary relationships and use the channel fusion module to conduct progressive context cue mining. \textbf{b) Transformer-based}: FSPNet \cite{FSPNet} leverages nonlocal token enhancement for improved feature interaction and utilizes a feature shrinkage decoder to aggregate camouflaged object cues better. RISNet \cite{RISNet} utilizes DepthGuided Feature Decoder merged RGB features and depth features, iterative feature refinement for improved small object detection. However, while these methods perform well on real images, their performance deteriorates on synthetic images. Therefore, we propose a new benchmark that combines domain adaptation from synthetic to real images with camouflaged object detection.\vspace{-5pt}

\subsection{Unsupervised Domain Adaptation}
The core idea of Unsupervised Domain Adaptation (UDA) is to improve performance on the target domain by learning from the source domain and narrowing the distributional gap between the source and target domain without requiring labeled data from the target domain. Among the various techniques developed for domain adaptation, three commonly explored approaches include divergence-based methods, adversarial learning, pseudo labeling and image-to-image translation with style transfer. \textbf{a) Divergence}: MMD \cite{MMD1, MMD2} is a statistical test used to determine if two distributions are equivalent based on their samples. CAN \cite{CAN} minimizes cross-entropy loss on labeled target data while alternating between estimating labels for target samples. \textbf{b) Adversarial Learning}: DANN \cite{DANN1, DANN2, DANN3} train the feature extractor by negating the gradient from the domain classifier using a gradient reversal layer during backpropagation. In ADDA \cite{ADDA}, the feature extractor is trained by inverting the labels for the domain classifier, making it learn domain-invariant representations. Shen et al. \cite{WDGRL} proposed WDGRL by modifying DANN, replacing the domain classifier with a network that directly approximates the Wasserstein distance.
\textbf{c) Pseudo Labeling}: Recently, teacher-student paradigms \cite{meanteacher} have been introduced to enable self-training for UDA, especially the source-free UDA (SFDA) task. For example, SSNLL~\cite{chen2022self} formulate SFDA task as a noisy label learning problem, and employ the teacher-student to tackle this problem. And some works~\cite{liu2023periodically, lin2023run} focus on the cooperation between the teacher and student models.
\textbf{d) Image-to-Image Translation and Style Transfer}: CycleGAN \cite{CycleGAN}, based on pix2pix, is a commonly used unsupervised image-to-image translation method, along with similar approaches like DualGAN \cite{DualGAN} and DiscoGAN \cite{DiscoGAN}. FDA \cite{FDA} uses Fast Fourier Transform to propose a simple method for domain alignment that does not require any learning. In addition to these classical techniques, recent efforts have started leveraging large-scale pre-trained vision-language models to improve cross-domain adaptation \cite{UDA-Clip, chen2025adapt}. Unlike the aforementioned methods, our approach explicitly frames UDA as a robust and efficient noisy label learning task, simplifying the challenge and delivering enhanced performance.

\begin{figure*}[ht]
    \centering
    \includegraphics[width=1.0\linewidth]{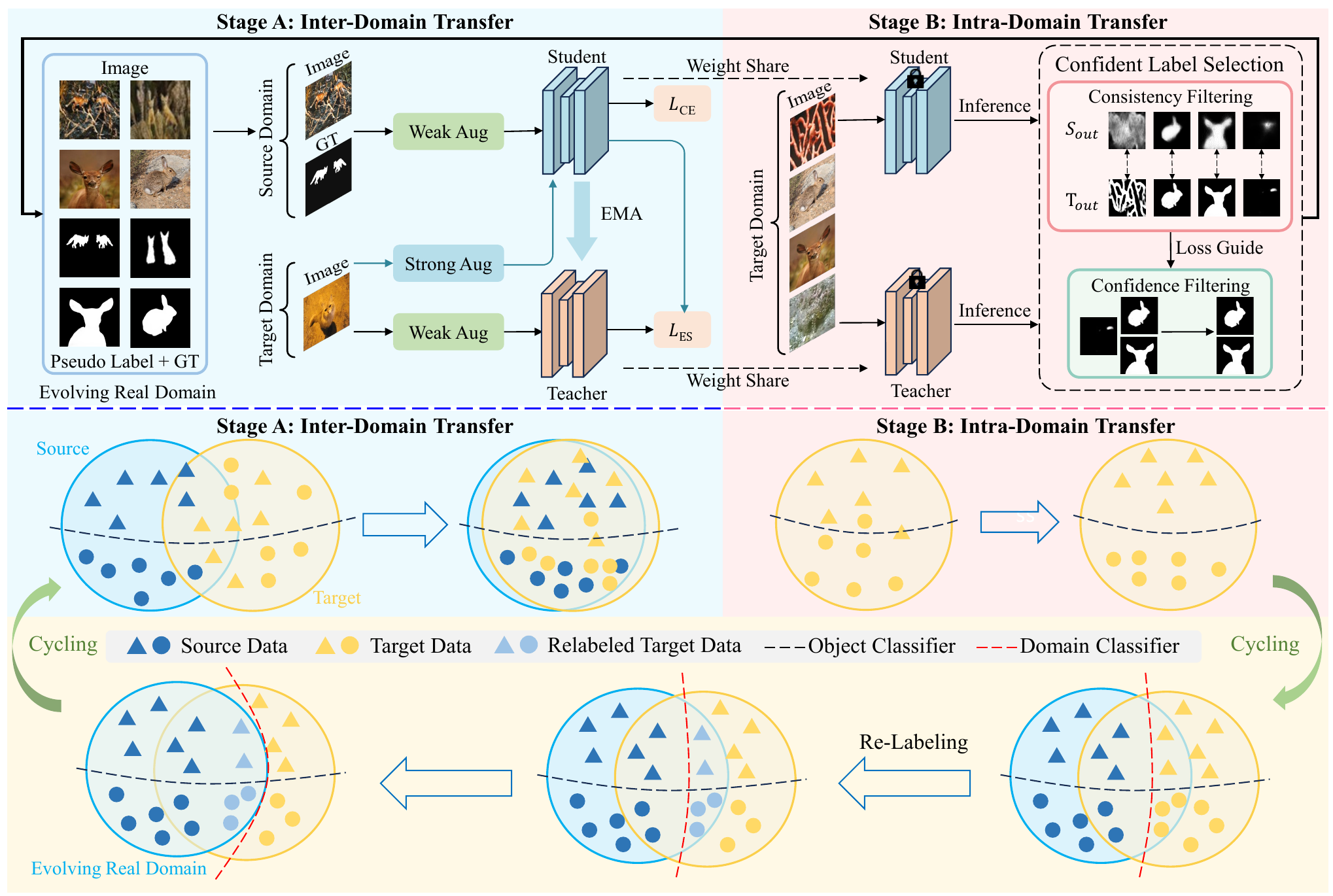}
    \caption{The overall pipeline of our CSRDA. 1) Stage A: EMA \cite{meanteacher} based student-teacher co-training with augmented synthetic source and real target data. 2) Stage B:  Confident Label Selection (CLS) evaluates the reliability of pseudo labels, and high confidence target domain samples are relabeled as source domain samples. Evolving Real Domain mechanism bridges domain gaps through a progressive domain alignment strategy, optimizing the cross domain feature space.}
    \label{fig:3}
\end{figure*}

\section{Method}
\subsection{Problem Setup}
From the perspective of unsupervised domain adaptation, we propose the first baseline model for S2R-COD. Given two domains, source domain $\mathcal{D}_{s}$ by generating models $g$ from auxiliary real datasets $\mathcal{D}_{r}$ synthesized to obtain:
\begin{equation}
    \left( x_s, y_s \right) = g \left( x_r, y_r \right), \ g \in \mathcal{G}_{COD-GEN},
\label{equ:test}
\end{equation}
where $(x_s, y_s) \in \mathcal{D}_s$ , and $(x_r,y_r) \in \mathcal{D}_r$, $x_s$ and $x_r$ denote source and origin real images, respectively, $y_s$, $y_r$ are their corresponding ground truths. $\mathcal{G}_{COD-GEN}$ is a method for generating synthetic COD images, such as LAKE-RED \cite{lake}, which embeds the ground truth $y_r$ in a random camouflaged background through a generative technique. The target domain $\mathcal{D}_{t}$ consists of unlabeled real camouflaged images. Note that $\mathcal{D}_{t}$ is distinct from $\mathcal{D}_{r}$ in terms of real images.
\begin{equation}
       P(\mathcal{D}_{t}) \neq P(\mathcal{D}_{r}),
\end{equation}
where $P$ denotes the data distribution. Our goal is to train a model $f:\mathcal{X}\rightarrow\mathcal{Y}$ using the domain adaptation method to use only the labeled data $\mathcal{D}_{s}$ and the unlabeled data $\mathcal{D}_{t}$, aiming to optimize the following objective:
\begin{equation}
\min\limits_{f}\mathbb{E}_{\left(x_s,y_s\right)\sim\mathcal{D}_{s},x_t\sim\mathcal{D}_{t}}\left[\mathcal{L}\left(f\left(x_s,x_t\right)\right),y_s\right],
\end{equation}
where $\mathcal{L}$ is total loss function and $y_t$ is unknown and inaccessible. The entire learning process is shown in \figref{fig:3}.
\subsection{Stage A: Inter-Domain Transfer}
We employ an EMA model $f_{ema}$ to generate pseudo labels inspired by the Mean Teacher \cite{meanteacher}, which has been widely used in self-supervised and semi-supervised learning \cite{bootstrap}. By leveraging temporal ensembling, EMA produces more reliable pseudo labels, improving their quality. In our framework, the teacher model $f_{ema}$ and the student model $f$ share the same backbone network (e.g., SINet \cite{SINet}) but take different image views as input, teacher model takes the weakly augmented image $\mathcal{A}\left(x_t\right)$, student model uses the weakly augmented image $\mathcal{A}\left(x_s\right)$ in the source domain and the strongly augmented image $\mathcal{\widetilde{A}}\left(x_t\right)$ in the target domain, During different training iterations, the teacher model is updated using the EMA of the student model. The weight of the teacher model can be expressed as follows:
\begin{equation}
    f_{ema}=\lambda f_{ema}+\left(1-\lambda\right)f,
\end{equation}
where $\lambda$ is the smoothing coefficient that controls the update momentum of teacher model. The optimization goal is defined as:
\begin{align}
\min\limits_{f}\ & \mathbb{E}_{(x_s, y_s) \sim \mathcal{D}_s, x_t \sim \mathcal{D}_t} \bigl[ \mathcal{L}_{CE} \bigl( f \bigl( \mathcal{A} \bigl( x_s \bigr) \bigr), y_s \bigr) \bigr] \notag \\
&+ \bigl[ \mathcal{L}_{ES} \bigl( f \bigl( \widetilde{\mathcal{A}} \bigl( x_t \bigr) \bigr), f_{ema} \bigl( \mathcal{A} \bigl( x_t \bigr) \bigr) \bigr) \bigr],
\end{align}

where $\mathcal{L}_{{CE}}$ is fully supervised loss of the student model using binary cross-entropy loss on source domain, and $\mathcal{L}_{{ES}}$ is customized loss function between the student model and teacher model.

\textbf{Edge-Aware Saliency-Weighted Loss.} To enhance the model ability to preserve edge structures and improve the learning of salient regions in cross domain camouflaged object detection, we designed an Edge-Aware Saliency-Weighted loss. This loss consists of two parts, the edge alignment loss and the significant region weighting loss. The total loss is formulated as:
\begin{equation}
    \mathcal{L}_{ES}=\alpha \mathcal{L}_{EA}+\beta \mathcal{L}_{SW},
\end{equation}
where $\alpha$ and $\beta$ are hyperparameters that balance the contributions of edge alignment and saliency weighted losses.
Edge alignment loss is defined as follows: 
\begin{equation}
    \mathcal{L}_{EA} = \mathbb{E}_{x_t \sim \mathcal{D}_t} \bigl\| \nabla f \bigl( \widetilde{\mathcal{A}} \bigl( x_t \bigr) \bigr) - \nabla f_{ema} \bigl( \mathcal{A} \bigl( x_t \bigr) \bigr) \bigr\|_1,
\end{equation}
where $\bigtriangledown$ represents the sobel edge operator, which is used to extract the edge intensity maps of the outputs from both the student and teacher models. The alignment of the edge structure between the two models is then enforced using the loss of L1. To strengthen the student model focus on salient regions, dynamic weights are assigned to the pseudo labels $\widehat{y}_t$ generated by the teacher model, and weighted cross-entropy is calculated:
\begin{align}
    \mathcal{L}_{S W} &= \mathbb{E}_{x_t \sim \mathcal{D}_t} \bigl\{ W \bigl[ \widehat{y}_t \log f\bigl(\mathcal{\widetilde{A}} \bigl(x_{t}\bigr)\bigr) \notag \\
    & \quad + \bigl(1-\widehat{y}_t\bigr) \log \bigl(1-f\bigl(\mathcal{\widetilde{A}}\bigl(x_{t}\bigr)\bigr)\bigr) \bigr] \bigr\}.
\end{align}

The weight vector $W=\widehat{y}_t+\delta$, where $\delta$ is a hyperparameter that boosts the weight of $\mathcal{L}_{SW}$. By reducing the weight of background or noise regions and increasing the weight of the prominent target regions, the interference of noise signals is mitigated while enhancing the model ability to fit key regions.

\begin{table*}[t!]
\captionsetup{skip=5pt}
\caption{Experimental results on HKU-IS $\rightarrow$ COD10K (\textbf{\boldmath$\hat{S}2C$}) benchmark. $\uparrow$ indicates the higher the score the better, and $\downarrow$ indicates the lower the better. The best results are marked in bold. The same representation is in the following tables.}
\centering
\renewcommand{\arraystretch}{1.0}
\setlength{\tabcolsep}{8.0pt}
\begin{tabular}{c|c|ccccccccc} 
\toprule
Model & Setting & $S_{\alpha }\uparrow$ & $F_{\beta }^{w}\uparrow$ & $E_{\phi }^{ad}\uparrow$ & $E_{\phi }^{mn}\uparrow$ & $E_{\phi }^{mx}\uparrow$ & $F_{\beta  }^{ad}\uparrow$ & $F_{\beta  }^{mn}\uparrow$ & $F_{\beta  }^{mx}\uparrow$ & $M\downarrow$ \\ 
\midrule
\multirow{3}{*}{SINet \cite{SINet}} & Source-Only & 0.6418 & 0.3707 & 0.7323 & 0.6570 & 0.7211 & 0.4653 & 0.4284 & 0.4718 & 0.0905 \\
& Mean Teacher & 0.6984 & 0.4165 & 0.7092 & 0.7177 & 0.7805 & 0.4888 & 0.5106 & 0.5598 & 0.0915 \\
& Ours & \textbf{0.7136} & \textbf{0.4814} & \textbf{0.7676} & \textbf{0.7443} & \textbf{0.7950} & \textbf{0.5548} & \textbf{0.5572} & \textbf{0.5960} & \textbf{0.0717} \\
\midrule
\multirow{3}{*}{SINet-v2 \cite{SINet-v2}} & Source-Only & 0.6466 & 0.4121 & 0.7332 & 0.6986 & 0.7035 & 0.4736 & 0.4640 & 0.4697 & 0.0976 \\
& Mean Teacher & 0.6485 & 0.4213 & 0.7143 & 0.7148 & 0.7347 & 0.4633 & 0.4709 & 0.4898 & 0.1115 \\
& Ours & \textbf{0.6845} & \textbf{0.4764} & \textbf{0.7707} & \textbf{0.7571} & \textbf{0.7643} & \textbf{0.5341} & \textbf{0.5319} & \textbf{0.5380} & \textbf{0.0787} \\
\midrule
\multirow{3}{*}{SegMaR \cite{SegMaR}} & Source-Only & 0.6468 & 0.4091 & 0.6947 & 0.6943 & 0.7199 & 0.4522 & 0.4648 & 0.4880 & 0.1215 \\
& Mean Teacher & 0.6597 & 0.4211 & 0.7040 & 0.7147 & 0.7382 & 0.4638 & 0.4783 & 0.4997 & 0.1071 \\
& Ours & \textbf{0.6832} & \textbf{0.4595} & \textbf{0.7298} & \textbf{0.7409} & \textbf{0.7619} & \textbf{0.5012} & \textbf{0.5190} & \textbf{0.5407} & \textbf{0.0787} \\
\midrule
\multirow{3}{*}{RISNet \cite{RISNet}} & Source-Only & 0.6627 & 0.4543 & 0.7332 & 0.7248 & 0.7310 & 0.5105 & 0.5059 & 0.5085 & 0.0752 \\
& Mean Teacher & 0.7622 & 0.5962 & 0.8173 & 0.8257 & 0.8400 & 0.6316 & 0.6502 & 0.6650 & 0.0529 \\
& Ours & \textbf{0.7627} & \textbf{0.6141} & \textbf{0.8301} & \textbf{0.8338} & \textbf{0.8401} & \textbf{0.6532} & \textbf{0.6621} & \textbf{0.6729} & \textbf{0.0516} \\
\bottomrule
\end{tabular}
\label{tab:1}
\end{table*}

\subsection{Stage B: Intra-Domain Transfer}
In the field of unsupervised image segmentation, the integration of noisy labels has emerged as a significant challenge. Since the generated noisy labels contain task related information cues, they can impose task specific regularization on the optimization process of unsupervised image segmentation. Inspired by the small-loss selection strategy in noisy label learning \cite{small-loss}, we propose a Confident Label Selection (CLS) mechanism to filter the pseudo labels generated by the teacher model in the target domain, extracting the cleaner portions and removing samples with high noise. Thus, we construct a high confidence pseudo label subset $\mathcal{D}_{cl}  \subseteq \mathcal{D}_t$ to reduce the impact of low quality pseudo labels on model training. Specifically, we leverage the prediction consistency between the teacher and student models in the target domain to assess the reliability of the pseudo labels and construct a high confidence pseudo label set based on the following criteria:
\begin{align}
\mathcal{D}_{cl} = \bigl\{ (x_t^{cl}, y_t^{cl}) \;\big|\,
&\mathcal{L}_{ES}(f(x_t^{cl}), f_{ema}(x_t^{cl})) \notag\\
& \le \mu \mathbb{E}_{x_t\sim \mathcal{D}_t } \mathcal{L}_{ES}(f(x_t), f_{ema}(x_t))\bigr\},
\end{align}
where $x_t^{cl} \in \mathcal{D}_t$ and $y_t^{cl}=f_{ema}(x_t^{cl})$, $\mu$ is a hyperparameter controls the selection ratio of pseudo labels, and the student and teacher models have fixed weights during this process. By dynamically selecting $\mathcal{D}_{cl}$ using the threshold $\mu$, an adaptive learning strategy \cite{pseudo} is essentially constructed. Furthermore, we impose a confidence constraint on the filtered pseudo labels, setting regions with a confidence lower than the threshold $\tau $ to zero and removing samples with overall confidence below $\tau $, ensuring that the model primarily learns high quality pseudo labels.
\begin{equation}
    \mathcal{D}_{cl} = \bigl\{ (x_t^{cl}, \widehat{y}_t^{cl}) \,\big|\, \widehat{y}_t^{cl} \ge \tau , \widehat{y}_t^{cl} \in {y}_t^{cl} \bigr\},
\end{equation}

 CLS mechanism guides the model to learn domain invariant features from clean pseudo labels, enhancing the representational power of the target domain and improving the generalization performance of the model. To reduce the domain gap, we introduce an evolving real domain $\widehat{\mathcal{D}}_{s}$ to bridge the source domain $\mathcal{D}_{s}$ and the target domain $\mathcal{D}_{t}$. Specifically, we replace $\mathcal{D}_{s}$ with $\widehat{\mathcal{D}}_{s}$ to obtain better pseudo labels. Our designs $\widehat{\mathcal{D}}_{s}$ through data fusion, leveraging the clear datasets $\mathcal{D}_{cl}$ from both the source and target domains.
\begin{equation}
    \widehat{\mathcal{D}}_{s}=\mathcal{D}_{s}\cup\mathcal{D}_{{cl}}.
\end{equation}
After obtaining $\widehat{\mathcal{D}}_{s}$, a new round of training is initiated using a student-teacher model. The final training objective of our framework can be expressed as follows:
\begin{align}
    &\min\limits_{f}\mathbb{E}_{(\widehat{x}_s, \widehat{y}_s) \sim \mathcal{\widehat{D}}_s, x_t \sim \mathcal{D}_t} [\mathcal{L}_{CE}\bigl(f\bigl(\mathcal{A}\bigl(\widehat{x}_s\bigr)\bigr), \widehat{y}_s\bigr) \nonumber \\
    &\quad + \mathcal{L}_{ES}\bigl(f\bigl(\widetilde{\mathcal{A}} \bigl(x_t\bigr)\bigr), f_{ema}\bigl(\mathcal{A}\bigl(x_t\bigr)\bigr)\bigr)].
    \label{eq:loss_function}
\end{align}

Through training, the model gradually learns the commonalities and differences between the source and target domain, thereby improving performance on the target domain. This approach not only effectively utilizes labeled source domain data and unlabeled target domain data but also mitigates the domain gap by synthesizing an evolving real domain, providing a novel solution for the unsupervised image segmentation problem.

\begin{table*}[t!]
\captionsetup{skip=5pt}
\caption{Experimental results on CAMO + NC4K + CHAM. $\rightarrow$ COD10K (\textbf{\boldmath$\hat{C}2C$}) benchmark.}
\centering
\renewcommand{\arraystretch}{1.0}
\setlength{\tabcolsep}{8.0pt}
\begin{tabular}{c|c|ccccccccc} 
\toprule
Model & Setting & $S_{\alpha }\uparrow$ & $F_{\beta }^{w}\uparrow$ & $E_{\phi }^{ad}\uparrow$ & $E_{\phi }^{mn}\uparrow$ & $E_{\phi }^{mx}\uparrow$ & $F_{\beta  }^{ad}\uparrow$ & $F_{\beta  }^{mn}\uparrow$ & $F_{\beta  }^{mx}\uparrow$ & $M\downarrow$ \\ 
\midrule
\multirow{3}{*}{SINet \cite{SINet}} & Source-Only & 0.6606 & 0.3999 & 0.6780 & 0.6807 & 0.7268 & 0.4471 & 0.4694 & 0.5074 & 0.1386 \\
 & Mean Teacher & 0.7299 & 0.4700 & 0.7225 & 0.7544 & 0.8179 & 0.5131 & 0.5607 & 0.5936 & 0.0860 \\
 & {Ours} & \textbf{0.7555} & \textbf{0.5202} & \textbf{0.7730} & \textbf{0.7779} & \textbf{0.8371} & \textbf{0.5679} & \textbf{0.6049} & \textbf{0.6297} & \textbf{0.0650} \\ 
\midrule
\multirow{3}{*}{SINet-v2 \cite{SINet-v2}} & Source-Only & 0.6753 & 0.4499 & 0.7135 & 0.7155 & 0.7519 & 0.4851 & 0.5047 & 0.5431 & 0.1235 \\
 & Mean Teacher & 0.6960 & 0.4958 & 0.7613 & 0.7673 & 0.7818 & 0.5300 & 0.5405 & 0.5624 & 0.0872 \\
 & {Ours} & \textbf{0.7211} & \textbf{0.5329} & \textbf{0.8020} & \textbf{0.7975} & \textbf{0.8069} & \textbf{0.5718} & \textbf{0.5817} & \textbf{0.5987} & \textbf{0.0669} \\ 
\midrule
\multirow{3}{*}{SegMaR \cite{SegMaR}} & Source-Only & 0.6670 & 0.4319 & 0.6997 & 0.7033 & 0.7445 & 0.4675 & 0.4843 & 0.5316 & 0.1231 \\
 & Mean Teacher & 0.7037 & 0.4860 & 0.7449 & 0.7574 & 0.7874 & 0.5170 & 0.5388 & 0.5724 & 0.0820 \\
 & {Ours} & \textbf{0.7312} & \textbf{0.5369} & \textbf{0.7959} & \textbf{0.7903} & \textbf{0.8065} & \textbf{0.5755} & \textbf{0.5894} & \textbf{0.6109} & \textbf{0.0660} \\ 
\midrule
\multirow{3}{*}{RISNet \cite{RISNet}} & Source-Only & 0.7192 & 0.5472 & 0.8054 & 0.7959 & 0.8015 & 0.5859 & 0.5875 & 0.6018 & 0.0702 \\
 & Mean Teacher & 0.7640 & 0.6006 & 0.8096 & 0.8108 & 0.8494 & 0.6088 & 0.6339 & 0.6844 & 0.0629 \\
 & {Ours} & \textbf{0.7797} & \textbf{0.6373} & \textbf{0.8263} & \textbf{0.8350} & \textbf{0.8603} & \textbf{0.6539} & \textbf{0.6647} & \textbf{0.6958} & \textbf{0.0507} \\ 
\bottomrule
\end{tabular}
\label{tab:2}
\end{table*}

\section{EXPERIMENTS}
\subsection{Settings}
\myPara{Dataset.} To validate the feasibility of the method, we performed evaluations on one SOD dataset and four COD datasets. HKU-IS \cite{HKU-IS} is a classic SOD benchmark dataset containing 4,447 images with complex backgrounds and multiple salient objects, covering both indoor and outdoor scenes. CAMO \cite{CAMO} consists of 1,250 camouflaged images and 1,250 non-camouflaged images. CHAMELEON \cite{CHAMELEON} includes 76 camouflaged images from natural scenes, focusing on animal camouflaged cases. COD10K \cite{SINet} is the largest COD dataset to date, containing 5,056 camouflaged images across five major categories and 69 subcategories. NC4K \cite{NC4K} includes 4,121 real world camouflaged object images. 

\myPara{Task Setup.} In this paper, we divide the S2R-COD tasks into \textbf{\boldmath$\hat{S}2C$} and \textbf{\boldmath$\hat{C}2C$}. It is important to note that, regardless of whether the dataset is SOD or COD, the corresponding synthetic COD dataset is used as the source domain. In \textbf{\boldmath$\hat{S}2C$} task, we use the synthetic HKU-IS dataset as the source domain, the real unlabeled COD10K training set as the target domain, and evaluate the performance on the COD10K test set. For \textbf{\boldmath$\hat{C}2C$} task, we consistently use the real unlabeled COD10K training set as the target domain and conduct final testing on different datasets. Based on the target test set, we adjust the source domain data selection as follows: 

\begin{table*}[t!]
\captionsetup{skip=5pt}
\caption{Experimental results on three additional \textbf{\boldmath$\hat{C}2C$} benchmarks, including (CAMO + NC4K $\rightarrow$ CHAM.), (CHAM. + NC4k $\rightarrow$ CAMO), and (CAMO + CHAM.  + COD10K $\rightarrow$ NC4K).}
\centering
\renewcommand{\arraystretch}{1.0}
\setlength{\tabcolsep}{6.8pt}
\begin{tabular}{c|c|cccccccccc}
\toprule
Test Dataset & Setting & $S_{\alpha }\uparrow$ & $F_{\beta }^{w}\uparrow$ & $E_{\phi }^{ad}\uparrow$ & $E_{\phi }^{mn}\uparrow$ & $E_{\phi }^{mx}\uparrow$ & $F_{\beta }^{ad}\uparrow$ & $F_{\beta }^{mn}\uparrow$ & $F_{\beta }^{mx}\uparrow$ & $M\downarrow$ \\ \midrule
\multicolumn{1}{c|}{\multirow{3}{*}{CHAMELEON \cite{CHAMELEON} }} & \multicolumn{1}{c|}{Source-Only} & 0.7063 & 0.5076 & 0.7939 & 0.7319 & 0.7773 & 0.5997 & 0.5711 & 0.6017 & 0.1189 \\
\multicolumn{1}{c|}{} & \multicolumn{1}{c|}{Mean Teacher} & 0.7667 & 0.5776 & 0.8287 & 0.7907 & 0.8360 & 0.6572 & 0.6489 & 0.6912 & 0.0927 \\
\multicolumn{1}{c|}{} & \multicolumn{1}{c|}{Ours} & \textbf{0.7801} & \textbf{0.6049} & \textbf{0.8558} & \textbf{0.8240} & \textbf{0.8731} & \textbf{0.6824} & \textbf{0.6707} & \textbf{0.7090} & \textbf{0.0715} \\ \midrule
\multicolumn{1}{c|}{\multirow{3}{*}{CAMO \cite{CAMO} }} & \multicolumn{1}{c|}{Source-Only} & 0.6387 & 0.4458 & 0.7139 & 0.6361 & 0.6643 & 0.5505 & 0.5138 & 0.5411 & 0.2002 \\
\multicolumn{1}{c|}{} & \multicolumn{1}{c|}{Mean Teacher} & 0.7055 & 0.5163 & 0.7872 & 0.7064 & 0.7545 & 0.6388 & 0.5896 & 0.6281 & 0.1414 \\
\multicolumn{1}{c|}{} & \multicolumn{1}{c|}{Ours} & \textbf{0.7158} & \textbf{0.5303} & \textbf{0.7994} & \textbf{0.7113} & \textbf{0.7781} & \textbf{0.6570} & \textbf{0.6021} & \textbf{0.6508} & \textbf{0.1301} \\ \midrule
\multicolumn{1}{c|}{\multirow{3}{*}{NC4K \cite{NC4K} }} & \multicolumn{1}{c|}{Source-Only} & 0.7554 & 0.5849 & 0.8128 & 0.7791 & 0.8224 & 0.6619 & 0.6541 & 0.6863 & 0.0926 \\
\multicolumn{1}{c|}{} & \multicolumn{1}{c|}{Mean Teacher} & 0.7887 & 0.5890 & 0.8271 & 0.7921 & 0.8594 & 0.6889 & 0.6848 & 0.7411 & 0.0880 \\
\multicolumn{1}{c|}{} & \multicolumn{1}{c|}{Ours} & \textbf{0.7946} & \textbf{0.6359} & \textbf{0.8528} & \textbf{0.7967} & \textbf{0.8698} & \textbf{0.7297} & \textbf{0.7076} & \textbf{0.7560} & \textbf{0.0737} \\ \bottomrule
\end{tabular}
\label{tab:3}
\end{table*}

\hspace{1em}\textit{Case 1:} When the test set is COD10K, the source domain consists of the remaining three synthetic COD datasets.

\hspace{1em}\textit{Case 2:} When the test set is CAMO or CHAMELEON, the source domain is composed of the synthetic COD datasets from Case 1 with CAMO or CHAMELEON images removed to prevent target domain data leakage. 

\hspace{1em}\textit{Case 3:} When the test set is NC4K since removing the corresponding synthetic data for NC4K leaves fewer source domain samples, we additionally introduce synthetic images from the COD10K test set to enhance the diversity of the source domain.

\myPara{Evaluation Metrics.} Referring to previous work \cite{ucos, camoteacher, GenSAM, UGTR, FEDER}, we use five metrics to comprehensively evaluate the model performance in COD task, including structure measure $(S_\alpha)$ \cite{S}, weighted F-measure $(F_{\beta }^{w})$ \cite{fw}, enhanced-alignment measure $(E_{\varphi })$ \cite{E1, E2}, F-measure $(F_{\beta })$ \cite{F}, and mean absolute error $(M)$. 

\myPara{Implementation Details.} All experiments are conducted on an RTX 3090 GPU and implemented using the PyTorch framework. We adopt SINet \cite{SINet} as the baseline of COD model and train student model using the Adam optimizer. The training process lasts for 40 epochs with a batch size of 16, initiating with a learning rate of 1e-4 and dividing it by 10 after 30 epochs. In \textbf{\boldmath$\hat{S}2C$} task, the smoothing coefficient $\lambda$ for teacher model is set to 0.996, while the hyperparameters $\mu$ and $\tau$ of CLS are set to 0.8 and 0.4, respectively. In contrast, for \textbf{\boldmath$\hat{C}2C$} task, these values are adjusted to 0.9996, 1.0, and 0.5, respectively. The number of cycling iterations is set to 2. During evaluation, only the teacher model is reserved for inference.

\begin{table*}
\captionsetup{skip=5pt}
\caption{Comparison of different UDA methods on CAMO + NC4K + CHAM. $\rightarrow$ COD10K (\textbf{\boldmath$\hat{C}2C$}) benchmark.}
\centering
\renewcommand{\arraystretch}{1.2}
\setlength{\tabcolsep}{10.0pt}
\begin{tabular}{c|ccccccccc}
\toprule
Method & $S_{\alpha }\uparrow$ & $F_{\beta }^{w}\uparrow$ & $E_{\phi }^{ad}\uparrow$ & $E_{\phi }^{mn}\uparrow$ & $E_{\phi }^{mx}\uparrow$ & $F_{\beta }^{ad}\uparrow$ & $F_{\beta }^{mn}\uparrow$ & $F_{\beta }^{mx}\uparrow$ & $M\downarrow$ \\ 
\midrule
Source-Only & 0.6606 & 0.3999 & 0.6780 & 0.6807 & 0.7268 & 0.4471 & 0.4694 & 0.5074 & 0.1386 \\ 
\midrule
FDA \cite{FDA} & 0.6847 & 0.4186 & 0.6896 & 0.7089 & 0.7618 & 0.4627 & 0.4885 & 0.5298 & 0.1064 \\ 
\midrule
DANN \cite{DANN3} & 0.7050 & 0.4572 & 0.7180 & 0.7382 & 0.7831 & 0.4939 & 0.5232 & 0.5571 & 0.0906 \\ 
\midrule
Mean Teacher \cite{meanteacher}  & 0.7299 & 0.4700 & 0.7225 & 0.7544 & 0.8179 & 0.5131 & 0.5607 & 0.6174 & 0.0873 \\ 
\midrule
PMT \cite{PMT}  & 0.6737 & 0.4861 & 0.7442 & 0.7423 & 0.7442 & 0.5227 & 0.5189 & 0.5205 & 0.0797 \\ 
\midrule
Ours & \textbf{0.7555} & \textbf{0.5202} & \textbf{0.7730} & \textbf{0.7779} & \textbf{0.8371} & \textbf{0.5679} & \textbf{0.6049} & \textbf{0.6530} & \textbf{0.0650} \\ 
\bottomrule
\end{tabular}
\label{tab:4}
\end{table*}

\subsection{Results and Analysis}
\myPara{Comparison with Baselines.}
In \tabref{tab:1} and \tabref{tab:2}, we compare the Source-Only baseline, Mean Teacher \cite{meanteacher}, and our method on \textbf{\boldmath$\hat{S}2C$} and \textbf{\boldmath$\hat{C}2C$} tasks. From the experimental results, we observe that Source-Only exhibits limited performance across all models, indicating a significant domain gap between synthetic and real images. Mean Teacher improves model performance to some extent by introducing target domain data. However, since this method does not filter the pseudo labels generated by the teacher model, a performance drop is observed on certain metrics for some models. For example, in \textbf{\boldmath$\hat{S}2C$} task, both SINet \cite{SINet} and SINet-v2 \cite{SINet-v2} under the Mean Teacher method show a certain decrease in MAE and $E_{\phi }^{ad}$ compared to the Source-Only baseline, with the $E_{\phi }^{ad}$ metric averaging a 2\% drop. This suggests that unfiltered pseudo labels may introduce noise, which negatively impacts the model performance. Our proposed method outperforms both Source-Only and Mean Teacher on all metrics in \textbf{\boldmath$\hat{S}2C$} and \textbf{\boldmath$\hat{C}2C$} tasks. For example, on SINet, $F_{\beta }^{w}$ increased by 11.1\% and 12\%, and $F_{\beta  }^{mx}$ increased by 12.5\% and 13.5\%. Compared to Mean Teacher, $F_{\beta }^{w}$ and $F_{\beta  }^{mx}$ also increased by 6.4\%, 5\%, and 3.6\%, 4.4\% respectively. By comparing the overall experimental data in \tabref{tab:1} and \tabref{tab:2}, it can be observed that the performance on \textbf{\boldmath$\hat{S}2C$} task is significantly lower than that on \textbf{\boldmath$\hat{C}2C$} tasks, highlighting a substantial domain gap between SOD and COD. However, it is worth noting that despite this domain gap, synthesizing a large amount of COD data from easily accessible SOD images remains a valuable research direction. Synthetic data provides an initial training basis for the model, especially in scenarios where the cost of labeling real COD data is high \cite{AGLNet, FindNet, BGNet, CubeNet, BSA-Net}, and can effectively alleviate the problem of data scarcity. As documented in \tabref{tab:3}, we conduct comprehensive evaluations on three mainstream COD datasets using SINet as the base model, validating the effectiveness of our proposed method. Compared with Source-Only and Mean Teacher, the proposed method presents a substantial advantage in terms of key performance indicators.

\begin{table}[]
\centering
\captionsetup{skip=5pt}
\caption{The ablation study of ES Loss and Confident Label Selection (CLS). MT denotes the Mean Teacher method.
}
\renewcommand{\arraystretch}{1.1}
\setlength{\tabcolsep}{1.3pt} 
\begin{tabular}{@{}ccccccc@{}}
\toprule
\multicolumn{7}{c}{\textbf{HKU-IS} $\rightarrow$ \textbf{COD10K} (\boldsymbol{$\hat{S}2C$})} \\ \midrule
\multicolumn{1}{c|}{Model} & \multicolumn{1}{c|}{Setting} & $S_{\alpha}\uparrow$ & $F_{\beta}^{w}\uparrow$ & $E_{\phi}^{mn}\uparrow$ & $F_{\beta}^{mn}\uparrow$ & $M\downarrow$ \\ \midrule
\multicolumn{1}{c|}{\multirow{3}{*}{SINet \cite{SINet}}} & \multicolumn{1}{c|}{MT+$\mathcal{L}_{CE}$} & 0.6984 & 0.4165 & 0.7177 & 0.5106 & 0.0915 \\
\multicolumn{1}{c|}{} & \multicolumn{1}{c|}{MT+$\mathcal{L}_{ES}$} & 0.7020 & 0.4531 & 0.7362 & 0.5284 & 0.0900 \\
\multicolumn{1}{c|}{} & \multicolumn{1}{c|}{MT+$\mathcal{L}_{ES}$+CLS} & \textbf{0.7136} & \textbf{0.4814} & \textbf{0.7433} & \textbf{0.5572} & \textbf{0.0717} \\ \midrule
\multicolumn{1}{c|}{\multirow{3}{*}{SINet-v2 \cite{SINet-v2}}} & \multicolumn{1}{c|}{MT+$\mathcal{L}_{CE}$} & 0.6485 & 0.4213 & 0.7148 & 0.4709 & 0.1115 \\
\multicolumn{1}{c|}{} & \multicolumn{1}{c|}{MT+$\mathcal{L}_{ES}$} & 0.6624 & 0.4426 & 0.7379 & 0.4928 & 0.0974 \\
\multicolumn{1}{c|}{} & \multicolumn{1}{c|}{MT+$\mathcal{L}_{ES}$+CLS} & \textbf{0.6845} & \textbf{0.4764} & \textbf{0.7571} & \textbf{0.5319} & \textbf{0.0787} \\ \midrule
\multicolumn{7}{c}{\textbf{CAMO + NC4K + CHAM.} $\rightarrow$ \textbf{COD10K} (\boldsymbol{$\hat{C}2C$})} \\ \midrule
\multicolumn{1}{c|}{Model} & \multicolumn{1}{c|}{Setting} & $S_{\alpha}\uparrow$ & $F_{\beta}^{w}\uparrow$ & $E_{\phi}^{mn}\uparrow$ & $F_{\beta}^{mn}\uparrow$ & $M\downarrow$ \\ \midrule
\multicolumn{1}{c|}{\multirow{3}{*}{SINet \cite{SINet}}} & \multicolumn{1}{c|}{MT+$\mathcal{L}_{CE}$} & 0.7299 & 0.4700 & 0.7544 & 0.5607 & 0.0873 \\
\multicolumn{1}{c|}{} & \multicolumn{1}{c|}{MT+$\mathcal{L}_{ES}$} & 0.7332 & 0.4812 & 0.7602 & 0.5658 & 0.0853 \\
\multicolumn{1}{c|}{} & \multicolumn{1}{c|}{MT+$\mathcal{L}_{ES}$+CLS} & \textbf{0.7555} & \textbf{0.5202} & \textbf{0.7779} & \textbf{0.6049} & \textbf{0.0650} \\ \midrule
\multicolumn{1}{c|}{\multirow{3}{*}{SINet-v2 \cite{SINet-v2}}} & \multicolumn{1}{c|}{MT+$\mathcal{L}_{CE}$} & 0.6960 & 0.4958 & 0.7673 & 0.5405 & 0.0872 \\
\multicolumn{1}{c|}{} & \multicolumn{1}{c|}{MT+$\mathcal{L}_{ES}$} & 0.7032 & 0.5015 & 0.7709 & 0.5475 & 0.0845 \\
\multicolumn{1}{c|}{} & \multicolumn{1}{c|}{MT+$\mathcal{L}_{ES}$+CLS} & \textbf{0.7211} & \textbf{0.5329} & \textbf{0.7975} & \textbf{0.5817} & \textbf{0.0669} \\ \bottomrule
\end{tabular}
\label{tab:5}
\vspace{-10pt}
\end{table}

\begin{figure}
    \centering
    \includegraphics[width=0.98\linewidth]{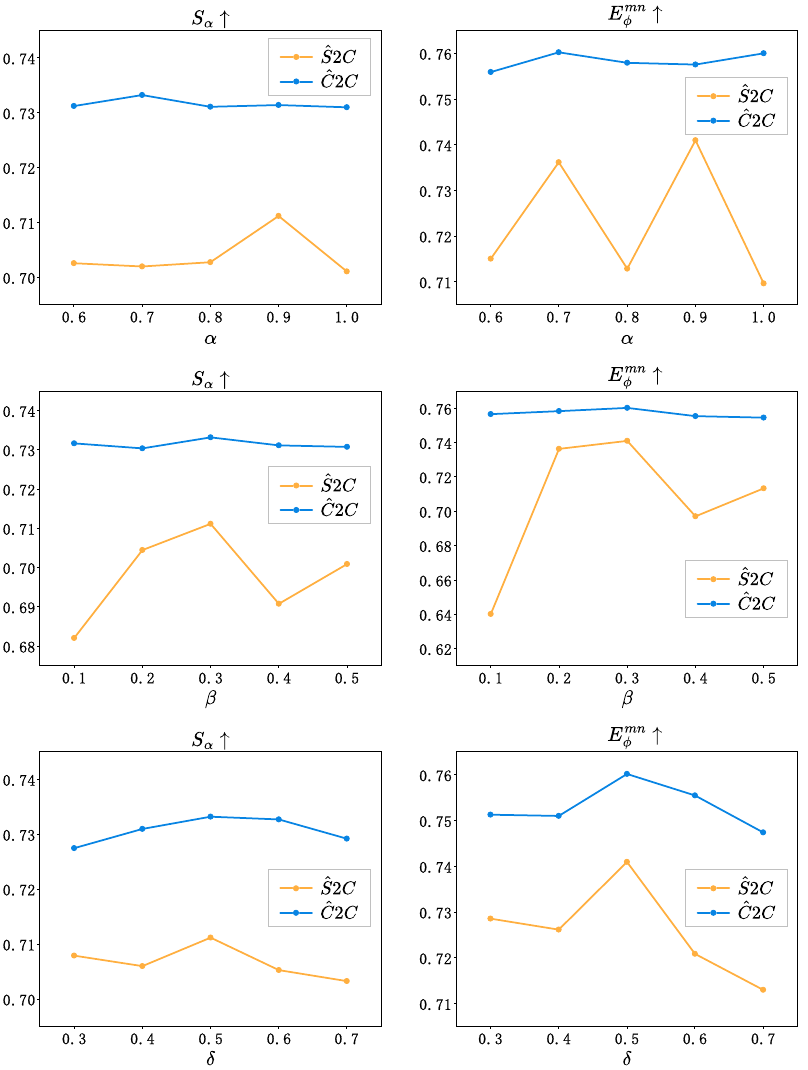}
    \caption{The impact of different hyperparameter combinations $\alpha$, $\beta$, $\delta$ on ES Loss under the HKU-IS $\rightarrow$ COD10K(\textbf{\boldmath$\hat{S}2C$}) and CAMO + NC4K + CHAM. $\rightarrow$ COD10K(\textbf{\boldmath$\hat{C}2C$}) benchmarks. The left figure presents the model performance in terms of $S_{\alpha}$, while the right figure shows the results for $E_{\phi}^{mn}$.}
    \label{fig:5}
    \vspace{-15pt}
\end{figure}

\myPara{Comparison with Other Related Methods.}
According to the experimental results on the COD10K benchmark for S2R-COD in \tabref{tab:4}, our method significantly outperforms existing UDA approaches across multiple critical metrics. Specifically, our approach achieves 0.7555 in comprehensive performance metric $S_{\alpha}$ and 0.5202 in $F_{\beta}^{w}$, surpassing the suboptimal Mean Teacher \cite{meanteacher} method by 2.5\% and 5.0\%, respectively. Compared to classical methods FDA \cite{FDA} and DANN \cite{DANN1}, our method demonstrates even greater improvements of 7.0\% and 5.0\% in $S_{\alpha}$. The robust performance on fine grained metrics, such as $E_{\phi }^{mx}$ (0.8371) and $F_{\beta }^{mx}$ (0.6530), highlights the model effectiveness in handling extreme target domain samples. Additionally, MAE (0.0650) indicates a 28.3\% reduction in false detection rates compared to DANN \cite{DANN1}, further validating our method superiority in domain adaptation.

\begin{figure*}[ht]
    \centering
    \includegraphics[width=0.98\linewidth]{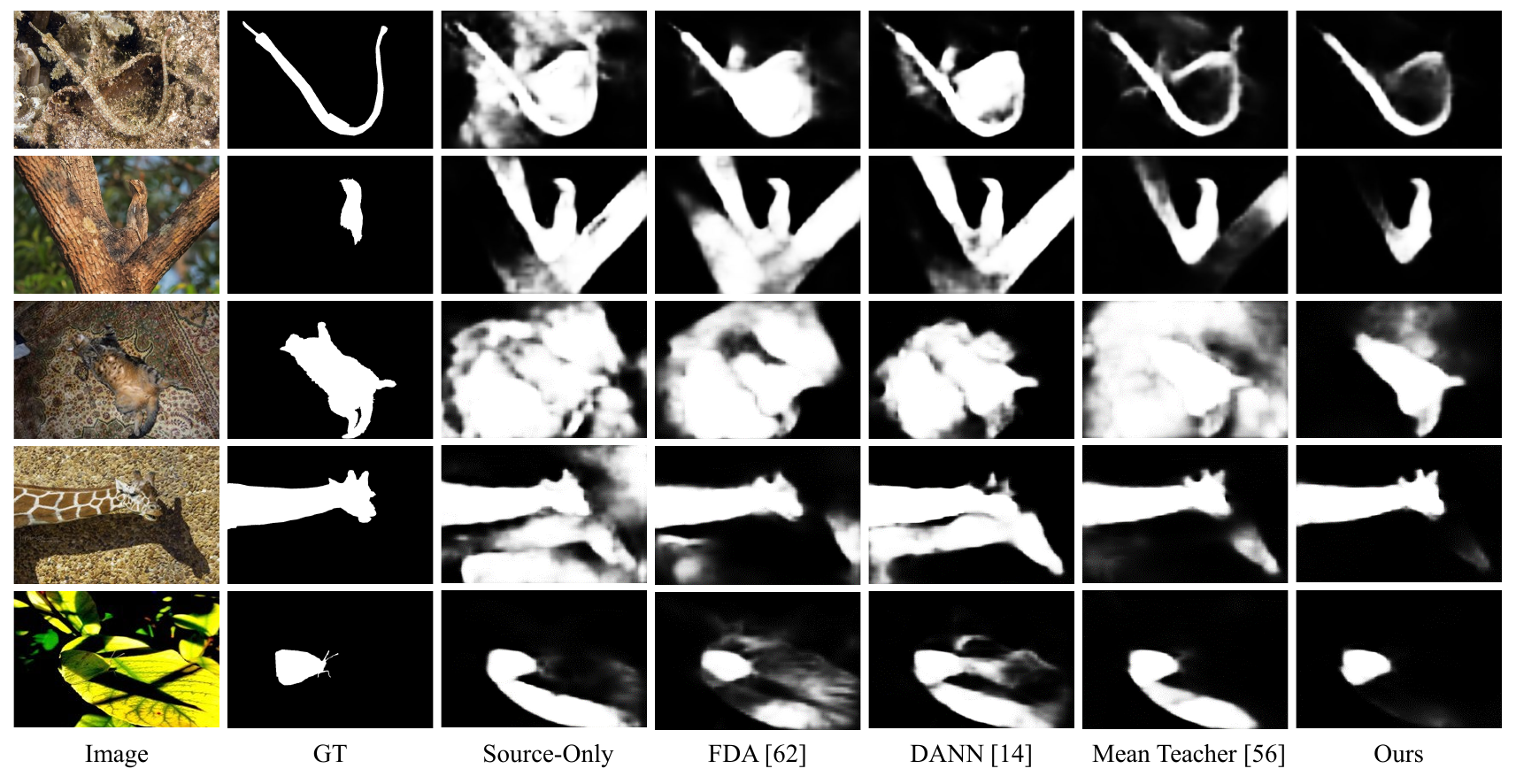}
    \vspace{-5pt}
    \caption{Visual comparisons with UDA methods on different types of samples. Please zoom in for details.}
    \label{fig:6}
    \vspace{-3pt}
\end{figure*}

\subsection{Ablation Studies}
\myPara{Effect of ES Loss and CLS.} The two key innovations in our work are ES loss and CLS. The ES Loss is specifically designed to enhance model robustness by improving its ability to handle camouflaged object boundaries and high confidence regions, thereby facilitating better adaptation to challenging target domains. Meanwhile, the CLS mechanism is introduced to refine the pseudo labels generated by the model, ensuring that cleaner and more reliable labels are retained in the source domain. This, in turn, mitigates the negative impact of noisy pseudo labels and helps to reduce the domain gap. To systematically evaluate the effectiveness of these two components, we conduct an ablation study and present the results in \tabref{tab:5}. The comparison across different model settings demonstrates that both the ES Loss and CLS contribute positively to performance improvements in both \textbf{\boldmath$\hat{S}2C$} and \textbf{\boldmath$\hat{C}2C$} tasks. Specifically, the ES loss provides more significant gains in \textbf{\boldmath$\hat{S}2C$} task, whereas the CLS brings greater improvements in \textbf{\boldmath$\hat{C}2C$} tasks. We hypothesize that in scenarios with a large domain gap (\textbf{\boldmath$\hat{S}2C$}), edge alignment and salient region weighting effectively enhance the model ability to capture target domain features. In contrast, for \textbf{\boldmath$\hat{C}2C$} tasks where the domain gap is smaller, the quality of pseudo labels plays a more critical role in determining model performance.

\myPara{Evaluation of Hyperparameters.} In \figref{fig:5}, we conduct a detailed analysis of the impact of hyperparameters $\alpha$, $\beta$ and $\delta$ in the ES Loss on the COD10K benchmark. Specifically, we examine how variations in these parameters influence the model performance in terms of $S_{\alpha}$ and $E_{\phi}^{mn}$. From the experimental results, we notably observe that for \textbf{\boldmath$\hat{S}2C$} task achieves its best performance when configured with $\alpha$ = 0.9, $\beta$ = 0.3, and $\delta$ = 0.5, which corresponds to the edge loss, weighted cross-entropy loss, and weight matrix, respectively. In contrast, for \textbf{\boldmath$\hat{C}2C$} task, the optimal choice of $\alpha$ differs, with $\alpha$ = 0.7 achieving better performance, while the other two parameters remain the same as in \textbf{\boldmath$\hat{S}2C$} task.

\subsection{Qualitative Evaluation}
\myPara{Visualization Analysis.} In \figref{fig:6}, we present the visualization results of the SINet \cite{SINet} on the COD10K dataset using different UDA methods. The Source-Only (Column 2) exhibits significant missegmentation due to domain differences, such as misidentifying branch textures as camouflaged objects. While the Mean Teacher \cite{meanteacher} method (Column 6) improves overall detection, it suffers from significant edge blurring. Other UDA methods also tend to mistakenly segment regions with similar characteristics around the target, a problem particularly evident in the fourth and fifth images, where shadows corresponding to camouflaged objects are frequently misidentified as part of the target. In contrast, our method (Column 7) effectively mitigates these issues by leveraging edge alignment constrained by ES Loss to achieve precise target contour localization in complex backgrounds (e.g., the butterfly in Row 5). Additionally, the CLS mechanism effectively suppresses artifactual noise (as seen in the background region of Row 4) while reducing the missegmentation of similar regions around the target. These improvements significantly enhance detection accuracy, further demonstrating the superior performance of our approach.

\vspace{-5pt}
\section{Conclusion}
This paper introduces a novel COD task, termed S2R-COD, which aims to leverage a set of easily accessible, annotated synthetic images to enhance model performance on real unlabeled target domain images. To achieve this, we propose CSRDA, a framework that employs a cycling student-teacher model to generate pseudo labels for real target domain images. Using CLS to filter these pseudo labels and introduce an evolving real domain to bridge the gap between the labeled source domain and the unlabeled target domain, thereby reducing domain discrepancies. Notably, we propose an innovative ES loss function to improve the model ability to capture target domain features. Furthermore, we establish a set of strong baselines to highlight the effectiveness of our approach. In summary, we hope this study provides new insights into synthetic-to-real camouflaged object detection.

\begin{acks}
This work was supported by the National Natural Science Foundation of China (Grant No. 62406071, Grant No. 62495061), and the China Postdoctoral Science Foundation(Grant No. 2025T180422, Grant No. 2024M761682, Grant No. GZB20240357) and Shui Mu Tsinghua Scholar (Grant No. 2024SM079).
\end{acks}

\bibliographystyle{ACM-Reference-Format}
\balance
\bibliography{sample-base}

\end{document}